%% file: main.tex
\documentclass[conference]{IEEEtran}

\usepackage{makeidx}
\usepackage{graphicx}
\usepackage{amsmath,amssymb} 
\usepackage{color}
\usepackage{graphicx}
\usepackage[utf8]{inputenc} 
\usepackage[T1]{fontenc}    
\usepackage{hyperref}       
\usepackage{url}            
\usepackage{booktabs}       
\usepackage{amsfonts}       
\usepackage{nicefrac}       
\usepackage{microtype}      
\usepackage{subcaption}
\usepackage{listings}
\usepackage{amsmath}
\usepackage{breakcites}
\usepackage{cleveref, amsfonts}

\DeclareMathOperator*{\argmax}{arg\,max}
\DeclareMathOperator*{\argmin}{arg\,min}

\begin{document}

\title{Adversarial Vulnerability of Active Transfer Learning}

\author{\IEEEauthorblockN{Nicolas M. Müller}
\IEEEauthorblockA{\textit{Cognitive Security Technologies} \\
\textit{Fraunhofer AISEC}\\
Garching near Munich \\
nicolas.mueller@aisec.fraunhofer.de}
\and
\IEEEauthorblockN{Konstantin Böttinger}
\IEEEauthorblockA{\textit{Cognitive Security Technologies} \\
\textit{Fraunhofer AISEC}\\
Garching near Munich \\
konstantin.boettinger@aisec.fraunhofer.de}
}
%

\maketitle

\begin{abstract}
\input{abstract}
\end{abstract}

\input{contents}
\bibliographystyle{acm}
\bibliography{bibliography} 

\end{document}

%% file: abstract.tex
Two widely used techniques for training supervised machine learning models on small datasets are Active Learning and Transfer Learning. 
The former helps to optimally use a limited budget to label new data. 
The latter uses large pre-trained models as feature extractors and enables the design of complex, non-linear models even on tiny datasets.
Combining these two approaches is an effective, state-of-the-art method when dealing with small datasets.

In this paper, we share an intriguing observation:
Namely, that the combination of these techniques is particularly susceptible to a new kind of data poisoning attack:
By adding small adversarial noise on the input, it is possible to create a collision in the output space of the transfer learner. 
As a result, Active Learning algorithms no longer select the optimal instances, but almost exclusively the ones injected by the attacker. 
This allows an attacker to manipulate the active learner to select and include arbitrary images into the data set, even against an overwhelming majority of unpoisoned samples. 
We show that a model trained on such a poisoned dataset has a
significantly deteriorated performance, dropping from 86\% to 34\% test accuracy.
We evaluate this attack on both audio and image datasets and support our findings empirically.
To the best of our knowledge, this weakness has not been described before in literature.

%% file: contents.tex
\section{Introduction}\label{s:intro}
Training supervised machine learning algorithms such as neural networks requires large amounts of labeled training data.
In order to solve problems for which there is no or little training data, previous work has developed techniques such as Transfer Learning and Active Learning.

Transfer Learning (TL) applies knowledge gained from one problem to a second problem. 
For example, consider the problem of training a Neural Network on a very small image dataset (say $N=500$).
Training directly on the dataset will yield poor generalization due to the limited number of training examples.
A better approach is to use a second, already-existing network trained on a different task to extract high-level semantic features from the training data, and train one's network on these features instead of the raw images themselves.
For images, this is commonly employed practice and easily accessible via the \emph{tensorflow} library. For audio data, similar feature extractors are also easily accessible ~\cite{pan2009survey}.

Active Learning (AL) on the other hand is a process where a learning algorithm can query a user.
AL is helpful when creating or expanding labeled datasets. Instead of randomly selecting instances to present to a human annotator, the model can query for specific instances.
The model ranks the unlabeled instances by certainty of prediction.
Those instances for which it is most uncertain are the ones where a label is queried from the human annotator.
Model uncertainty can be understood as the distance to the decision surface (SVM), or entropy of the class predictions (\emph{uncertainty sampling} for Neural Networks).
In summary, Active Learning can help in finding the optimal set of instances to label in order to optimally use a given budget~\cite{settles2009active}.

Since these approaches are complementary, they can be combined straightforwardly:
One designs a \emph{transfer active learning} system by combining an untrained network with a pre-trained feature extractor and then allows this combined model to query a human expert.
Previous work has examined this in detail and finds that it can accelerate the learning process significantly~\cite{al_tl,al_tl_b,al_tl_c}.

In this work, we are the first to observe that this combination of AL with TL is highly susceptible to data poisoning.
Our contribution is to present this novel weakness, which
\begin{itemize}
    \itemsep0em
    \item allows an attacker to reliably control the samples queried by the active learner (94.8 \% success rate even when the poisoned samples are outnumbered 1:50 by clean training data),
    \item considerably deteriorates the test performance of the learner (by more than 50 percent in absolute test accuracy),
    \item is hard to detect for a human annotator.
\end{itemize}
We evaluate the attack on both audio and image datasets and report the results in Section~\ref{s:results}.
To the best of our knowledge, this attack has not been described in literature before whatsoever.

\section{Related Work}\label{s:related_work}
In this section, we discuss related work on transfer active learning.
We first present work on combining transfer and active learning and then discuss related data poisoning attacks for each approach. 
We observe that there is no prior work on data poisoning for the combined method of transfer active learning.
\subsubsection{Active Learning with Transfer Learning.}\label{sss:related_work_al_tl}
Kale et Al. \cite{al_tl} present a transfer active learning framework which "leverages pre-existing labeled data from related tasks to improve the performance of an active learner".
They take large, well-known datasets such as \emph{Twenty Newsgroup}, and evaluate the number of queries required in order to reach an error of $0.2$ or less.
They can reduce the number of queries by as much as 50\% by combining transfer learning with active learning.
The authors of \cite{al_tl_b} perform similar experiments, and find that "the number of labeled examples required for learning with transfer is often significantly smaller than that required for learning each target independently".
They also evaluate combining active learning and transfer learning, and find that the "combined active transfer learning algorithm [...] achieve[s] better prediction performance than alternative methods".

Chan et al. \cite{chan-ng-2007-domain} examine the problem of training a word sense disambiguation (WSD) system on one domain and then applying it to another domain, thus 'transferring' knowledge from one domain to another domain. 
They show that active learning approaches can be used to help in this transfer learning task.
The authors of \cite{shi_domain_transfer} examine how to borrow information from one domain in order to label data in another domain.
Their goal is to label samples in the current domain (in-domain) using a model trained on samples from the other domain (out-of-domain).
The model then predicts the labels of the in-domain data.
Where the prediction confidence of the model is low, a human annotator is asked to provide the label, otherwise, the model's prediction is used to label the data.
The authors report that their approach significantly improves test accuracy.

\subsubsection{Related Data Poisoning Attacks.}\label{sss:related_work_poisoning}
Data poisoning is an attack on machine learning systems where malicious data is introduced into the training set. When the model trains on the resulting poisoned training dataset, this induces undesirable behavior at test time.
Biggio et Al. \cite{biggio2012poisoning} were one of the first to examine the effects of data poisoning on Support Vector Classifiers.
The branch of data poisoning most related to our work is \emph{clean poison} attacks.
These introduce minimally perturbed instances with the 'correct' label into the data set.
For example, the authors of \cite{shafahi2018poison} present an attack on transfer learners which uses \emph{clean poison} samples to introduce a back-door into the model.
The resulting model misclassifies the targeted instance, while model accuracy for other samples remains unchanged.
To give an example, these \emph{clean poison} samples may be manipulated pictures of dogs that, when trained on by a transfer learner, will at test time cause a \emph{specific} instance of a cat to be classified as a dog.
Other samples than the targeted 'cat' instance will not be affected.
Such \emph{clean label} attacks have also been explored in a black-box scenario~\cite{zhu2019transferable}.

Data Poisoning not only affects classification models but also regression learners.
Jagielski et Al. \cite{jagielski2018manipulating} present attacks and defenses on regression models that cause a denial of service:
By injecting a small number of malicious data into the training set, the authors induce a significant change in the prediction error.

\subsubsection{Poisoning Active Learning.}\label{sss:related_work_al_poisoning}
Poisoning active learners requires the attacker to craft samples which, in addition to adversely affecting the model, have to be selected by the active learner for labeling and insertion into the dataset. This attack aims at both increasing overall classification error during test time as well as increasing the cost for labeling.
In this sense, poisoning active learning is harder than poisoning conventional learners, since two objectives have to be satisfied simultaneously.
Miller et Al. \cite{miller2014adversarial} present such an attack: They poison linear classifiers and manage to satisfy both previously mentioned objectives (albeit with some constraints):
Poisoned instances are selected by the active learner with high probability, and the model trained on the poisoned instances induces significant deterioration in prediction accuracy.

\subsubsection{Adversarial Collisions}
There exists some related work on adversarial collisions, albeit with a different focus from ours:
Li et Al.~\cite{li2019approximate} observe that neural networks can be insensitive
to adversarial noise of large magnitude. This results in 'two very different examples sharing the same feature activation' and results in a feature collision.
However, this constitutes an attack at test time (evasion attack), whereas we present an attack at training time (poison attack).

\section{Attacking Active Transfer Learning}
In this section, we introduce the proposed attack.
We first present 
the threat model and then detail the attack itself. 
In Section~\ref{s:results}, we evaluate the effectiveness of our attack empirically.


\subsection{Threat model}
In this work, we assume that an attacker has the following capabilities:
\begin{itemize}
    \item The attacker may introduce a small number of adversarially perturbed instances into the active learning pool.
    These instances are unlabeled. They are screened by the active learning algorithm and may, along with the benign instances, be presented to the human annotator for labeling.
    \item The attacker cannot compromise the output of the human annotator, i.e. they cannot falsify the labels assigned to either the benign or poison instances.
    \item The attacker knows the feature extractor used. This could, for example, be a popular open-source model such as \emph{resnet50}~\cite{he2016deep} or \emph{YAMNet}~\cite{Yamnet}.
    These models are readily available and widely used~\cite{ResNetan6:online}.
    \item The attacker has no knowledge about, or access to the model trained on top of the feature extractor.
    \item The attacker does not know the active learning algorithm.
\end{itemize}

\subsection{Feature Collision Attack}\label{ss:poisoning_collision_attack}
Since transfer active learning is designed to use found data, i.e. data from untrusted sources, it is highly susceptible to data poisoning attacks.
In this section, we present such an attack and show how it completely breaks the learned model.

Let $X, Y$ be a set of unpoisoned data, where the data $X$ and targets $Y$ consist of $N$ instances.
An instance pertains to one of $M$ different classes (e.g. 'dog' or 'cat').
Let $f$ be the pretrained feature extractor, which maps a sample $x_i \in X$ to a $d_\zeta$ dimensional feature vector (i.e. $f(x_i) = \zeta_i)$.
Let $g$ be the dense model head, which maps a feature vector $\zeta_i$ to some prediction $y_{pred} \in [0, 1]^{M}$, where $y_{pred}$ is a one-hot vector, i.e. $\sum_{m=0}^{M-1} y^m_{pred} = 1$.
Thus, an image is classified by the subsequent application of the feature extractor $f$ and the dense model head $g$:
\begin{equation}
    y = \argmax_i g(f(x_i))
\end{equation}

For a set of instances $\{x_i\}$, a set of adversarial examples \{$x_i + \delta_i$\} can be found by minimizing, for each $x_i$ separately:
\begin{equation}\label{eq:opt}
    \delta_i = \argmin_\delta \big \lVert f(x_i + \delta) - \mu \big \rVert_2 + \beta \lVert \delta \rVert_2
\end{equation}
where $\beta \in \mathbb{R}^+$ and $\mu$ is a fixed vector of size $d_\zeta$.
Solving Equation~\ref{eq:opt} will find adversarial examples that 1) are selected by the active learner 2) break the model, and 3) are imperceptible to the human annotator:
\begin{enumerate}
    \item \textbf{Examples are queried.} A set of thusly found adversarial examples $\{x_i + \delta_i\}$ will all be mapped to the same output $\mu$, i.e. $f(x_i + \delta_i) = \mu$ for all $i$.
    This will 'confuse' the active learner, since all adversarial examples share the same feature vector, but have different class labels.
    Thus, the active learner will incur high uncertainty for instances mapped to this \emph{collision vector} $\mu$, and thus will query almost exclusively these (we verify this experimentally in Section~\ref{ss:results}).
    \item \textbf{Examples are harmful}. These examples will break any model head trained on the extracted features $\zeta_i$, since all adversarial examples share identical features, but different labels.
    \item \textbf{Examples are undetected by a human annotator.} 
    Once queried, the adversarial examples $x_i + \delta_i$ will be reviewed by a human annotator and labeled accordingly.
    The second part of Equation~\ref{eq:opt} ensures that the adversarial noise is small enough in magnitude to remain undetected by human experts. The adversarial example will thus be assigned the label of $x_i$, but not raise suspicion of the human annotator. The scalar $\beta$ is a hyperparameter controlling the strength of this regularisation.
\end{enumerate}


\subsubsection{Choice of Collision Vector}
The \emph{Collision Vector} $\mu$ is chosen as the zero vector $\mu = 0^{d_\zeta}$ because of two reasons:
First, we find that it helps numerical convergence of Equation~\ref{eq:opt} when training with Gradient Descent.
Second, a zero-vector of features is highly uncommon with unpoisoned data.
Thus, it induces high uncertainty with the active learner, which in turn helps to promote the adversarial poison samples for labeling and inclusion in the training dataset from the beginning.
It is possible to chose a different $\mu$, for example the one-vector $\mu=1^{d_\zeta}$, or the mean of the feature values $\mu = \overline{\zeta_i}$. However, we find that the zero-vector works best, most likely due to the reasons detailed above.

\subsubsection{Improving attack efficiency}
We propose two improvements over the baseline attack.
First, when choosing the base instances from the test set to poison and to include in the train set,
it is advisable to select those where
\begin{equation}
    \lVert f(x) - \mu \rVert_2
\end{equation}
is smallest.
Intuitively, this pre-screens the samples for those where the optimization step (Equation~\ref{eq:opt}) requires the least work.
Second, maintaining class balance within the poison samples improves effectiveness.
This helps in maximally confusing the active learner since a greater diversity of different labels for the same feature vector $\mu$ increases the learner's uncertainty with respect to future samples that map to $\mu$.
In all the following analysis, we evaluate the attack with these improvements in place.

\section{Implementation and Results}\label{s:results}

In this section, we first describe our transfer active learning setup and the data sets used for evaluation.
We then implement our attack and demonstrate its effectiveness (c.f. Section~\ref{ss:results}).

\subsection{Active Transfer Learner Setup}
This section describes the data we use and our choice of transfer learner.

For our experiments, we use image and audio data.
This is because Active Learning requires a human oracle to annotate instances, and humans are very good at annotating both image and audio data, but rather inefficient in processing purely numerical data.
This motivates the choice of the active learner, namely a neural network, which in recent times has been shown to provide state-of-the-art performance on image and audio data.

Thus, we create the transfer learner as follows:
We use a large, pre-trained model to perform feature extraction on the audio and image data.
For image data, we use a pre-trained \emph{resnet50} model~\cite{he2016deep}, which comes with the current release of the python \emph{tensorflow} library.
For audio data, we build a feature extractor from the \emph{YAMNet} model, a deep convolutional audio classification network~\cite{Yamnet}.
Both of these feature extractors map the raw input data to a vector of features.
For example, \emph{resnet50} maps images with $32*32*3 = 3072$ input dimensions to a vector of $2048$ higher-level features.
A dense neural network (\emph{dense head}) is then used to classify these feature vectors.
Our Active Learner uses Entropy Sampling~\cite{settles2009active} to compute the uncertainty
\begin{align}
    \text{uncertainty($x_i$)} &= H(g(f(x_i))) \\ &= - \sum_{m \in M} g(f(x_i))^m \log g(f(x_i))^m
\end{align}
for all unlabeled $x_i$. The scalar value $g(f(x_i))^m$ indicates softmax-probability of the $m$-th class for of the network's output.
The active learner computes the uncertainty for all unlabeled instances $x_i$ and selects the one with the highest uncertainty to be labeled.

\subsubsection{Prevention of Overfitting.}\label{sss:overfitting}
As detailed in Section~\ref{s:intro}, we use active transfer learning in order to learn from very small datasets.
Accordingly, we use at most $N=500$ instances per dataset in our experiments.
In this scenario, overfitting can easily occur.
Thus, we take the following countermeasures:
First, we keep the number of trainable parameters low and use a dense head with at most two layers and a small number of hidden neurons.
Second, we use high Dropout (up to $50\%$) and employ early stopping during training.
Third, we refrain from training the weights of the Transfer Learner (this is commonly referred to as \emph{fine-tuning}).
This is motivated by the observation that the \emph{resnet50} architecture has more than 25 Million trainable weights, which makes it prone to overfitting, especially on very small datasets.

\subsubsection{Datasets}\label{sss:datasets}
We use three datasets to demonstrate our attack. 
\begin{itemize}
    \itemsep0em
    \item \emph{Google AudioSet}~\cite{AudioSet_2017}, a large classification dataset comprising several hundred different sounds. We use only a subset of ten basic commands (\emph{right}, \emph{yes}, \emph{left}, etc.).
    \item \emph{Cifar10}~\cite{cifar10}, a popular image recognition dataset with 32x32 pixel images of cars, ships, airplanes, etc.
    \item \emph{STL10}~\cite{STL-10_dataset_2011}, a semi-labeled dataset, comprising several thousand labeled 96x96 pixel images in addition to tens of thousand unlabeled images, divided into ten classes. In this supervised scenario, we use only the labeled subset.
\end{itemize}
Each dataset is split into a \emph{train} set and two test sets, \emph{test1} and \emph{test2} using an 80, 10 and 10 percent split.
Following previous work~\cite{shafahi2018poison}, we train on the \emph{train} set, use the \emph{test1} set to craft the adversarial examples, and evaluate the test accuracy on the \emph{test2} set. 
Thus, the adversarial examples base instances originate from the same distribution as the train images, but the two sets remain separate.

\subsection{Feature Collision Results}\label{ss:results}
\begin{figure}[h]
    \centering
    \includegraphics[trim={0 5cm 0 5cm},clip, width=0.49\linewidth]{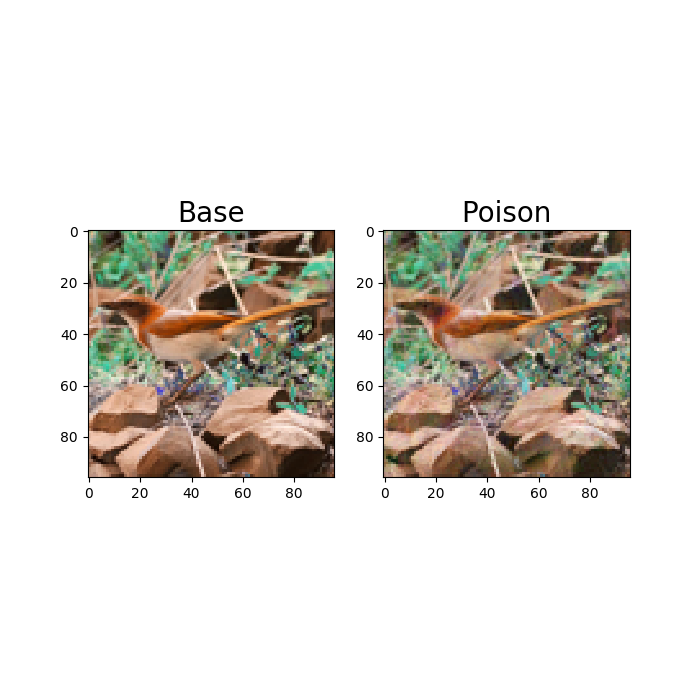}
    \includegraphics[trim={0 5cm 0 5cm},clip, width=0.49\linewidth]{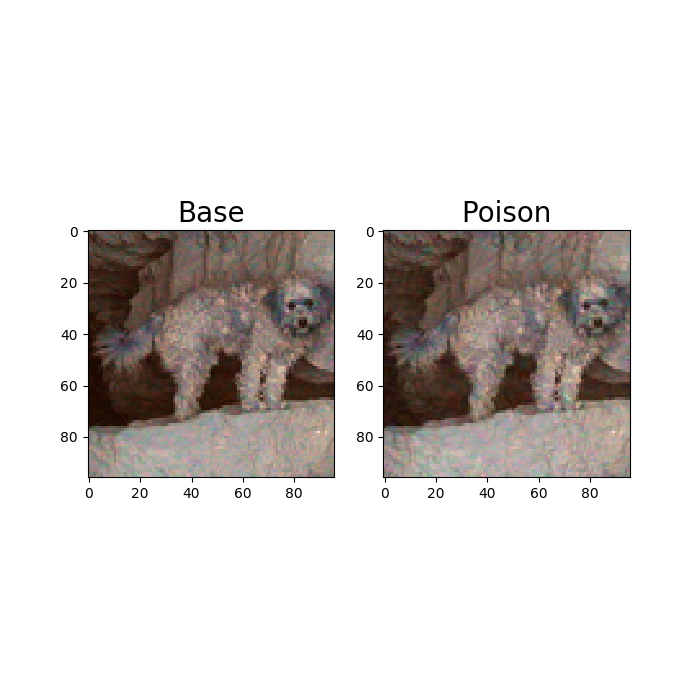}
    \includegraphics[trim={0 5cm 0 5cm},clip, width=0.49\linewidth]{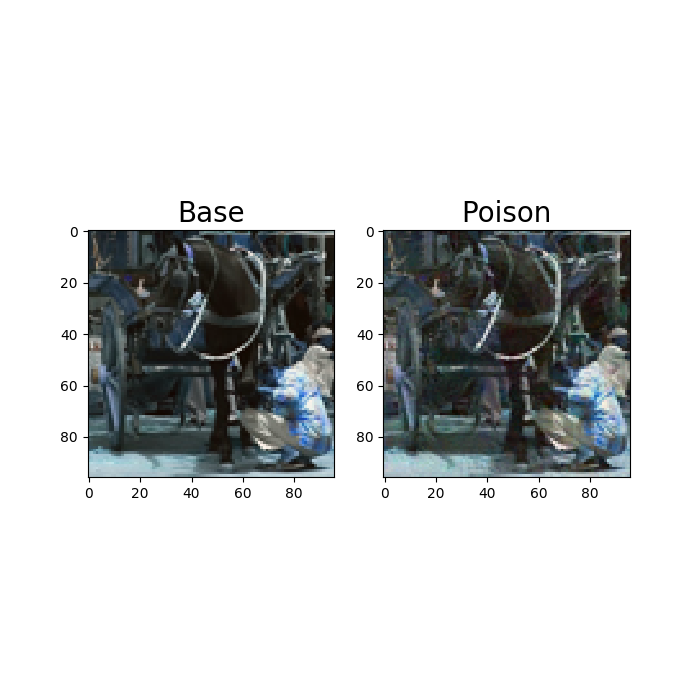}
    \includegraphics[trim={0 5cm 0 5cm},clip, width=0.49\linewidth]{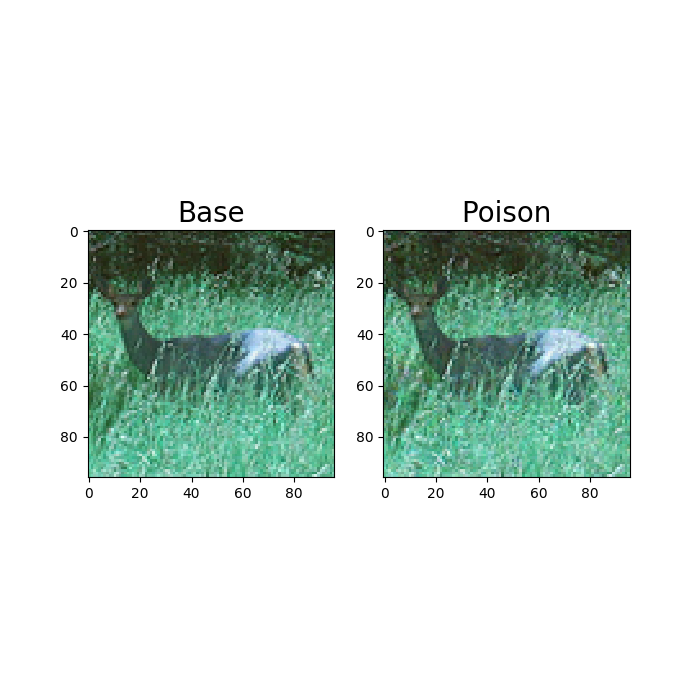}
    \caption{Images with index 155 (bird), 308 (horse), 614 (dog) and 3964 (deer) from the STL dataset. The left image of each pair shows the base instance, i.e. the unpoisoned image. The right image shows the poisoned image which causes the collision in the transfer learner's output space.}
    \label{fig:poisons}
\end{figure}

We now verify empirically that the created instances look inconspicuous and are hard to distinguish from real, benign instances.
Consider Figure~\ref{fig:poisons}.
It shows four randomly selected poison instances of the $STL10$ dataset.
Observe that the poisoned images are hardly distinguishable from the originals.
We also provide the complete set of audio samples for the Google Audio Set\footnote{\url{https://drive.google.com/file/d/1JtXUu6degxnQ84Kggav8rgm9ktMhyAq0/view}}, both poisoned and original .wav files.

\begin{figure}
    \centering
    \includegraphics[width=0.8\linewidth]{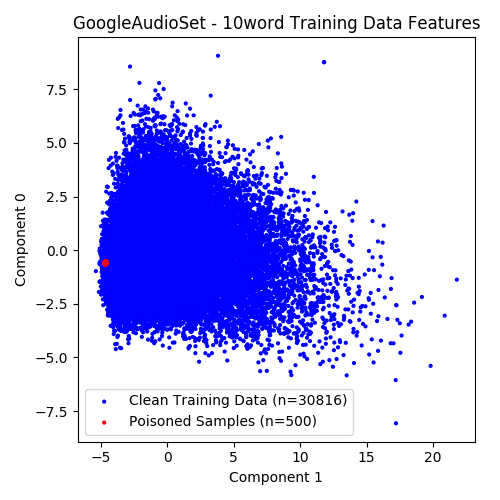}
    \includegraphics[width=0.8\linewidth]{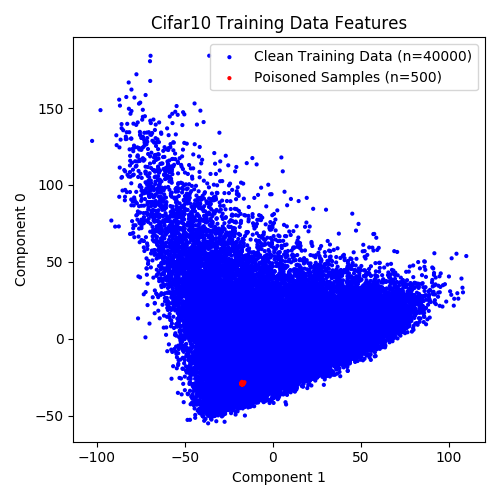}
    \caption{Visualisation of the transfer learner's feature space. The top image shows the first two principal components of the \emph{Google AudioSet 10} dataset's features when using the YAMNet feature extractor.
    The bottom image shows the same visualization of the \emph{Cifar10} dataset, using the \emph{imagenet50} feature extractor.
    The blue dots represent the unpoisoned training data. 
    The red dots represent poison data found via Equation~\ref{eq:opt}, chosen equally per class (50 instances for each of the ten classes).
    Observe the large diversity of the unpoisoned training data compared to the adversarial poison data, which is projected onto a single point, thus creating a 'feature collision' which massively deteriorates the active learner's performance.
    }
    \label{fig:feature_collisions}
\end{figure}

We now proceed to visually illustrate the results of the feature collision.
Consider Figure~\ref{fig:feature_collisions}:
For the poisoned and benign training data in the \emph{AudioSet10} dataset, it shows the corresponding feature vectors after PCA-projection in two dimensions. 
Thus, it shows what the dense model head 'sees' when looking at the poisoned data set.
Note that while the unpoisoned data has a large variety along the first two principal components (as is expected, since the individual instances pertain to different classes), the poison data's features collide in a single point (red dot) - even though they also pertain to different classes.
Thus, we observe a 'feature collision' of the poison samples to a single point in feature space, which, in combination with the different labels, will cause maximum 'confusion' in the active learner.

\subsection{Impact on the Model}
In this section, we evaluate the impact on the classification accuracy of transfer active learners when exposed to the poison samples.
For each of the three datasets, we create 500 poison samples and include them into the training set.
We then create a transfer active learner (a neural network with one / two layers), train it on 20 unpoisoned samples, and simulate human annotation by letting the active learner query for 500 new instances from the training set (which contains a majority of benign data plus the injected poison samples).
We find that the active learner 
\begin{itemize}
    \itemsep0em
    \item almost exclusively selects poisoned samples, and
    \item test performance is degraded severely (by up to 50\% absolute).
\end{itemize}
Table 1 details our results.

\input{result_table.tex}

For example, consider the first row, which evaluates a one-layer neural network (NN1) on the STL-dataset.
In an unpoisoned scenario, after having queried 500 images, the active learner has a test-accuracy of 86 percent.
When introducing 500 poison instances, we observe the following:
First, even though the poison instances are outnumbered 1:8, the active learner chooses 500 out of 500 possible poison instances for manual annotation - a success rate of $100$ percent.
In comparison, if the adversarial instances would be queried with the same probability as unpoisoned samples, only $12.6$ percent of them would be chosen (random success rate).
Secondly, observe that test accuracy is degraded significantly from 86 percent to 34 percent.
This is because the model can not learn on data that, due to the feature collision attack, looks identical to one another for the dense head.

\subsection{Hyper Parameters and Runtime}
Table 1 details the run time to find a single adversarial example on an Intel Core i7-6600 CPU (no GPU), which ranges from one to two minutes.
We used the following hyperparameters to find the adversarial samples: For the audio samples, we used $\beta = 0.3$ and $2000$ iterations with early stopping and adaptive learning rate.
For the image samples, we used $\beta = 1e-5$ and $500$ iterations.
The difference in $\beta$ between image and audio data is due to the difference in input feature scale, which ranges from $\pm 127$ for images to $\pm 2^{15}$ for audio data.\footnote{16bit audio has a feature range of $\pm[0,..., 2^{15}-1]$, where one bit is reserved for the sign.}


\subsection{Adversarial Retraining Defense}
We find that there exists a trivial, yet highly effective defense against our proposed attack:
Unfreezing, i.e. training of the feature extractor $f$.
When $f$ is trained in conjunction with $g$, training on the poison samples actually serves as 'adversarial retraining', which boosts model robustness~\cite{adv_retrain}.
In our experiments, we find that restraining completely negates the effects of the poisoning attack.
However, it is not a satisfactory defense due to the following concerns.
\begin{itemize}
    \item \textbf{Lack of labeled samples}. In order to unfreeze the weights of the feature extractor $f$, a large number of labeled samples is required. These are not available at the start of the active learning cycle. Thus, retraining can only occur during later stages. Until then, however, the adversary has free reign to introduce their poison samples into the dataset.
    \item \textbf{High computational overhead}. Retraining the feature extractor $f$ incurs high computational overhead in comparison to training only the dense head $g$.
    \item \textbf{Overfitting}. Training $f$, especially on a small dataset, may result in overfitting, as described in Section~\ref{sss:overfitting}.
\end{itemize}

In summary, while unfreezing and adversarial retraining does mitigate the attack, these strategies may be hard to apply due to several practical concerns.
Thus, a better defense strategy is required, which we leave for future work.

\section{Conclusion and Future Work}
In this work, we point out an intriguing weakness of the combination of active and transfer learning:
By crafting feature collisions, we manage to introduce attacker-chosen adversarial samples into the dataset with a success rate of up to 100 percent.
Additionally, we decreased the model's test accuracy by a significant margin (from 86 to 36 percent).
This attack can effectively break a model, wastes resources of the human oracle, and is very hard to detect for a human when reviewing the poisoned instances.
To the best of our knowledge, this particular weakness of transfer active learning has not been observed before.

%% file: result_table.tex
\begin{table*}[]
\caption{Results of the feature collision poisoning on three data sets: Cifar10, STL and Google Audio Set. The Head of the Active Learner has one (NN1) or two (NN2) dense layers. The success rate provides the ratio with with poison samples are selected by the active learner.}
\centering
\resizebox{\textwidth}{!}{%
\begin{tabular}{|c|c|c|c|c|c|c|c|c|c|}
\hline
Dataset           & Model & \begin{tabular}[c]{@{}c@{}}Accuracy\\ (Clean)\end{tabular} & \begin{tabular}[c]{@{}c@{}}Accuracy \\ (Poisoned)\end{tabular} & \begin{tabular}[c]{@{}c@{}}Loss \\ (adv.)\end{tabular} & \begin{tabular}[c]{@{}c@{}}Loss \\ (initital)\end{tabular} & N     & \begin{tabular}[c]{@{}c@{}}Success \\ Rate \\ (Poison)\end{tabular} & \begin{tabular}[c]{@{}c@{}}Success \\ Rate \\ (Random)\end{tabular} & \begin{tabular}[c]{@{}c@{}}Time \\ (s)\end{tabular} \\ \hline
STL                   & NN1   & 0.862                                                      & 0.347                                                          & 107.729                                                & 66450.624                                                  & 500 & 1.0                                                              & 0.126                                                            & 113.08                                               \\\hline
Cifar10                 & NN1   & 0.432                                                      & 0.267                                                          & 39.746                                                 & 5016.588                                                   & 500 & 1.0                                                              & 0.013                                                            & 93.425                                               \\\hline
Audio Set 10 & NN1   & 0.252                                                      & 0.143                                                          & 0.488                                                  & 24.59                                                      & 500 & 1.0                                                              & 0.016                                                            & 73.083                                               \\\hline
STL                   & NN2   & 0.844                                                      & 0.7                                                            & 107.729                                                & 66450.624                                                  & 500 & 0.86                                                             & 0.126                                                            & 113.08                                               \\\hline
Cifar10                 & NN2   & 0.428                                                      & 0.341                                                          & 39.746                                                 & 5016.588                                                   & 500 & 0.832                                                            & 0.013                                                            & 93.425                                               \\\hline
Audio Set 10 & NN2   & 0.263                                                      & 0.141                                                          & 0.488                                                  & 24.59                                                      & 500 & 0.998                                                            & 0.016                                                            & 73.083                                              \\\hline
\end{tabular}%
}
\end{table*}